\documentclass{article}

\usepackage[dvipsnames]{xcolor}         %
\usepackage{iclr2024_conference}
\iclrfinalcopy

\usepackage{amsmath,amsfonts,bm}

\def\eqref#1{equation~\ref{#1}}

\def\1{\bm{1}}

\DeclareMathAlphabet{\mathsfit}{\encodingdefault}{\sfdefault}{m}{sl}
\SetMathAlphabet{\mathsfit}{bold}{\encodingdefault}{\sfdefault}{bx}{n}

\newcommand{\E}{\mathbb{E}}

\usepackage[colorlinks,citecolor=blue]{hyperref}       %
\usepackage{url}

\usepackage{algorithm,algorithmicx}
\usepackage{wrapfig}
\usepackage[noend]{algpseudocode}
\usepackage{times}
\usepackage[utf8]{inputenc} %
\usepackage[T1]{fontenc}    %
\usepackage{booktabs}       %
\usepackage{amsfonts}       %
\usepackage{nicefrac}       %
\usepackage{microtype}      %

\usepackage{graphicx}
\usepackage{subfigure}
\usepackage{multirow}
\usepackage{enumerate}
\usepackage{amsthm,amsmath,amssymb}
\usepackage{dsfont}
\usepackage{mathtools}
\usepackage[noend]{algpseudocode}
\usepackage{xspace}

\usepackage[capitalize,noabbrev]{cleveref}

\usepackage{listings}

\PassOptionsToPackage{numbers}{natbib}
\usepackage{natbib}

\usepackage{std_definitions}

\newcounter{qcounter}

\usepackage{ifthen}

\newcommand{\dfc}[1]{}

\algnewcommand{\LeftComment}[1]{\Statex \(\triangleright\) #1}
\algdef{SE}[SUBALG]{Indent}{EndIndent}{}{\algorithmicend\ }%
\algtext*{Indent}
\algtext*{EndIndent}

\newcommand{\ppo}{\texttt{PPO}}
\newcommand{\cpi}{$\texttt{PPO}^{++}$}
\newcommand{\agg}{\texttt{AggreVaTeD}}
\newcommand{\lol}{\texttt{LOLS}}
\newcommand{\lolsd}{\texttt{D}$^2$\texttt{LOLS}}

\newcommand{\bc}{\texttt{SFT}}
\newcommand{\bestofn}{\texttt{Best-of-N}}

\newcommand{\bcppo}{\texttt{SFT}+\texttt{PPO}}
\newcommand{\bccpi}{\texttt{SFT}+$\texttt{PPO}^{++}$}
\newcommand{\bcagg}{\texttt{SFT}+\texttt{AggreVaTeD}}
\newcommand{\bclol}{\texttt{SFT}+\texttt{LOLS}}
\newcommand{\bclolsd}{\texttt{SFT}+\texttt{D}$^2$\texttt{LOLS}}

\newcommand{\gpt}{\texttt{GPT2}}

\newcommand{\framework}{\mbox{RLGF}}

\definecolor{forestgreen2}{rgb}{0.13, 0.55, 0.13}

\newcommand{\piref}{\pi^{g}}

\title{Learning to Generate Better Than Your LLM}

\author{%
  Jonathan D. Chang\thanks{Equal contribution} \\
  Department of Computer Science\\
  Cornell University\\
  \texttt{jdc396@cornell.edu} \\
  \And
  $\text{Kiant\'e  Brantley}^*$ \\
  Department of Computer Science\\
  Cornell University\\
  \texttt{kdb82@cornell.edu} \\
  \AND
  Rajkumar Ramamurthy \\
  Fraunhofer IAIS\\
  \texttt{rajkumar.ramamurthy@iais.fraunhofer.de} \\
  \And
  Dipendra Misra \\
  Microsoft Research New York\\
  \texttt{dipendra.misra@microsoft.com} \\
  \And
  Wen Sun \\
  Department of Computer Science\\
  Cornell University\\
  \texttt{ws455@cornell.edu} \\
}

\begin{document}

\maketitle

\begin{abstract}

Reinforcement learning (RL) has emerged as a powerful paradigm for fine-tuning Large Language Models (LLMs) for text generation. In particular, recent LLMs such as ChatGPT and GPT-4 can engage in fluent conversations with users after finetuning with RL. %
Capitalizing on key properties of text generation, we seek to investigate RL algorithms beyond general purpose algorithms like Proximal Policy Optimization (PPO). In particular, we extend RL algorithms to allow them to interact with a dynamic black-box guide LLM and propose RL \emph{with guided feedback} (\framework{}), a suite of RL algorithms for LLM fine-tuning. We provide two ways for the guide LLM to interact with the LLM  to be optimized for maximizing rewards. The guide LLM can generate text which serves as additional starting states for the RL optimization procedure. The guide LLM can also be used to complete the partial sentences generated by the LLM that is being optimized, treating the guide LLM as an expert to imitate and surpass eventually.
We experiment on the IMDB positive sentiment, CommonGen, and TL;DR summarization tasks. We show that our RL algorithms achieve higher performance than supervised learning (SL) and the RL baseline PPO, demonstrating the benefit of interaction with the guide LLM. On both CommonGen and TL;DR, we not only outperform our SL baselines but also improve upon PPO across a variety of metrics beyond the one we optimized for. Our code can be found at \url{https://github.com/Cornell-RL/tril}.
\end{abstract}

\section{Introduction}
Large Language Models (LLMs) have become very capable in various real-world applications ranging from being able to answer open-ended questions on numerous topics~\citep{zhang2022greaselm}, write articles from short descriptions~\citep{goyal2022news}, generate code~\citep{Githubcopilot}, follow robot commands~\citep{huanginner}, solve puzzles~\citep{bubeck2023sparks}, and even showcased as assistive models for education~\citep{KhanAcademy} and healthcare~\citep{lee2023benefits}. 

However, using supervised learning (SL) to train LLMs presents a challenging metric mismatch \citep{wiseman2016sequence} between the training and testing regimes. The metric mismatch arises from the training metric being the log-loss while the testing metrics are task-specific such as BLEU or user satisfaction rating. This discrepancy is magnified when fine-tuning LLMs on downstream tasks where the main goal is not just producing fluent text but also being proficient at solving the specific task. Another mismatch is the training and testing distributions mismatch. SL methods train model on the given static datasets, while in inference time, the LLMs need to make prediction \emph{conditioned on the text it has generated by itself}. Such a distribution mismatch during training and testing has been widely observed in literature such as Imitation Learning and RL \citep{ross2011reduction}, robotics \citep{ross2013learning}, and  NLP \citep{bengio2015scheduled, arora2022exposure}.\looseness=-1

Reinforcement Learning (RL) addresses these mismatches by directly optimizing the metrics through reward feedback \emph{on the states generated by the RL agent itself}. The ability to test in real world and obtain reward feedback to correct and improve the agents' behaviors on the fly makes RL a more powerful learning paradigm than SL. 
Recently, OpenAI fine-tuned LLMs with RL from human feedback (RLHF) to better align LLMs to human intentions, leading to the great success of ChatGPT \citep{chatgpt}. Following this, multiple other models trained with RL such as Anthropic's Claude2 \citep{claude2} and Meta's LLama2 \citep{touvron2023llama} further proved the effectiveness of RL. Recently, GRUE benchmark \citep{ramamurthy2022rl4lm} systematically studied RL versus SL when finetuning LLMs on downstream tasks with predefined rewards. GRUE's preliminary results demonstrate the benefit of RL when fine-tuning LLMs, leading to the release of popular codebases such as RL4LMs \citep{ramamurthy2022rl4lm}, TRLx \citep{trlx} and AlpacaFarm \citep{dubois2023alpacafarm}, that enables RL for language models. However, ChatGPT, RL4LMs, TRLX, and AlpacaFarm all use vanilla policy gradient methods known to be sample inefficient and sensitive to local minima due to the combinatorially large search space of natural language generation \citep{ramamurthy2022rl4lm}. \looseness=-1

\begin{figure*}
    \vspace{-2mm}
    \includegraphics[width=\textwidth]{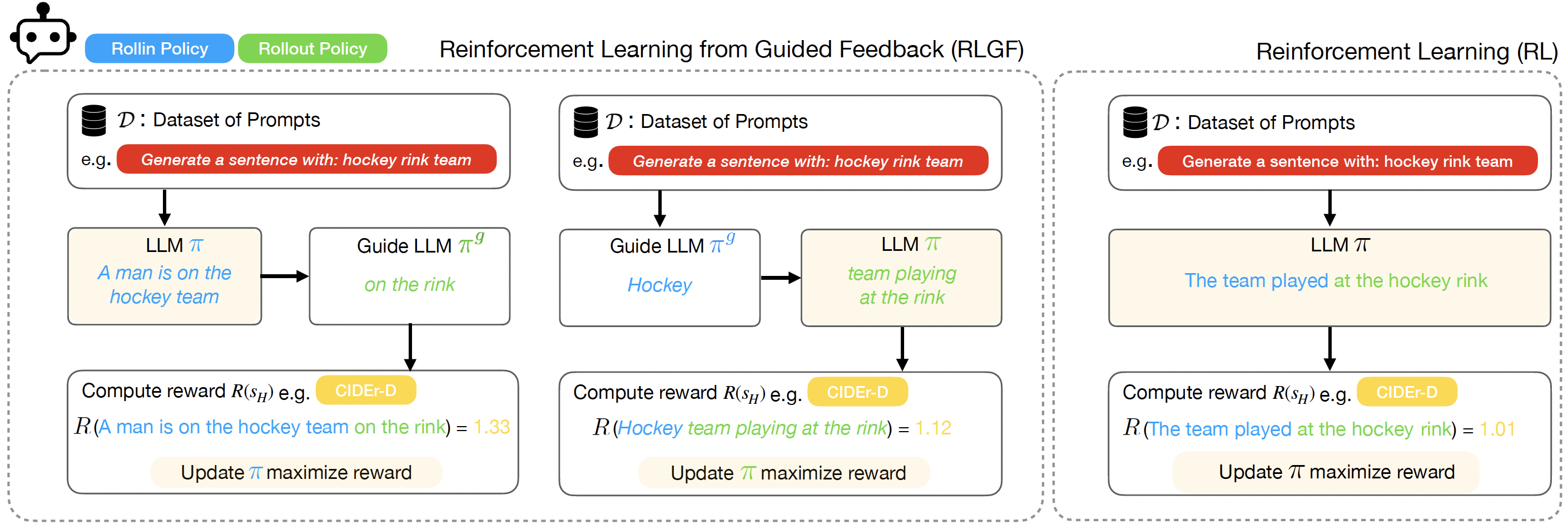}
    \vspace{-1mm}
    \caption{(Left) Reinforcement Learning with Guided Feedback (RLGF) flow chart showing how breaking up generations into two parts, rollins and rollouts done by different LLMs opens up a rich framework of interaction when compared to (Right) Reinforcement Learning (RL). RLGF uses a guide policy $\piref$ to guide the policy training on $\pi$ for maximizing a reward function. The guide policy $\piref$ can be used to complete the partial sentences generated by the policy $\pi$, which allows the RL algorithms to treat $\piref$ as an expert to imitate and surpass. \framework{} can also use $\piref$ to generate partial sentences from which the RL algorithms start optimizing $\pi$. 
    RLGF treats $\piref$ as a black-box model which gives RLGF the flexibility of using different pre-trained LLMs (or even a human expert) as $\piref$.  We mainly experiment using a supervised fine tuned model followed by nucleus sampling (\texttt{SFT+nucleus}) as $\piref$. Our experiments show that RLGF is capable of learning a policy that is better than $\piref$ and the policy learned by standard RL alone.
    }
    \vspace{-1mm}
    \label{fig:main-flowchart}
\end{figure*}

In this work, we focus on more efficient RL methods for fine-tuning LLMs on downstream tasks with predefined rewards (e.g., well-defined metric such as Bleu, or reward learned from human preference feedback). Our approach is motivated by the classic prior work on \emph{RL with rich reset distributions} \citep{kakade2002cpi,bagnell2003policy} and \emph{Imitation Learning (IL)} \citep{ross2011reduction,sun2017aggrevated, chang2015learning}, which often leverages an existing guide policy (not necessarily an optimal policy) to reduce the search space for more efficient and optimal learning. Our key observation is that since modern pre-trained LLMs exhibit impressive general language capabilities, they can serve as guide policies to improve the RL procedure. Our framework, which we call, \emph{RL with guided feedback} ($\framework$), integrates a guide policy into a policy gradient framework (Fig.~\ref{fig:main-flowchart}). When the guide policy can provide reasonable but potentially sub-optimal predictions for downstream tasks, our framework can then leverage to learn a near-optimal strategy. We introduce simple and novel algorithms for fine-tuning LLMs using our $\framework$ framework while capturing various existing IL and RL algorithms.  Our proposed algorithms are simple and introduce little overhead on computation and memory compared to PPO (especially when using LoRA adapters), making it straightforward to replace PPO by our algorithms in any RLHF pipeline. 

We evaluate on three tasks. The first is IMDB where the goal is to generate a positive and fluent review given an initial context. The second is CommonGen where the goal is to write a fluent text that uses a given set of words. Finally, we test on the TL;DR summarization task where the objective is to learn to generate summaries using human preference data. For all tasks, we find evidence of metric mismatch from SL-based fine-tuning approaches and show that RL-based methods which utilize reward signals outperform on the task metric. We then demonstrate \framework{} outperforming PPO on reward, fluency, as well as automated lexical metrics such as Rouge. In our experiments, our guide policy is the SFT model equipped with nucleus sampling. Thus comparing to the baseline PPO which uses the SFT model as a warm start, our algorithms use the same amount of information and thus is a fair comparison to PPO. Finally, we investigate how various baselines and \framework{} algorithms balance the inherent trade-off between reward optimization and the KL constraint in the RLHF objective. We provide both theoretical justification and empirical evidence to show the benefit of using RLGF for fine-tuning LLMs on downstream tasks. 

\section{Related Work}
Here we present the most relevant works at the intersection of IL, RL, and natural language generation. Please see Appendix \ref{appendix:related_work_full} for a more thorough treatment of the literature.

\textbf{IL for Structured Prediction: } 
Algorithms such as  Schedule Sampling (SS) \citep{bengio2015scheduled}, methods using SS \citep{duckworth2019parallel, mihaylova2019scheduled, goyal2017differentiable}, SEARNN \citep{leblond2017searnn}, Bridging the Gap \citep{zhang2019bridging}, Mixer \citep{ranzato2015sequence} been inspired by IL for structured prediction algorithms DAGGER \citep{ross2011reduction}, DAD \citep{venkatraman2015improving}, and SEARN \citep{daume2009search}. Our work is inspired by AggreVaTeD \citep{sun2017aggrevated} (Differentiable  AggreVaTe \cite{ross2014aggrevate})  where the algorithm makes use of differentiable policies and multi-step feedback rather than immediate one-step predictions to imitate. Similarly, we present a differentiable version of LOLS \citep{chang2015learning} as well as an improvement, D$^2$LOLS. \looseness=-1

\textbf{LLM Fine-tuning from Human Preferences: }
Recent advancements in fine-tuning of Large Language Models (LLMs) have shown incredible success in tasks through learning from human preferences. Being simpler to accumulate human preferences, Reinforcement Learning from Human Feedback (RLHF) \citep{stiennon2020learning} introduced a paradigm to utilize RL to improve downstream performance on translation \citep{kreutzer2018reliability}, summarization \citep{stiennon2020learning}, storytelling \citep{ziegler2019fine}, and instruction following \citep{chatgpt}. %
Another family of work use supervised learning style methods for fine-tuning LLMs \citep{zhao2023slic,yuan2023rrhf,rafailov2023dpo, liu2023statistical}. %
DPO, SLiC, RRHF, and RSO are methods that optimize for compatibility with a preference dataset under a preference reward model (either explicitly modeling a reward function or implicitly representing a reward function via an LLM itself) such as the Bradley Terry model \citep{bradley1952rank}. Whether or not one should use RL or SFT to fine-tune LLM is not the question we aim to address here, instead, our work mainly focus on improving PPO for fine-tuning LLMs, and our key contribution is novel RL algorithms that can outperform PPO on various tasks.
\looseness=-1

\textbf{LLM Distillation: }
With an ever growing arsenal of powerful, black-box LLMs, recent work has aimed to distill specific capabilities into a smaller model. Knowledge distillation \citep{bucilua2006model, hinton2015distilling} in autoregressive models investigated matching sequence level log probabilities \citep{kim2016sequence}, model hidden states \citep{jiao2019tinybert}, or attention scores \citep{wang2020minilm}. Recently, more sophisticated methods, inspired from the IL literature, are being proposed to better imitate the expert LLM's performance \citep{lin2020autoregressive, agarwal2023gkd, mukherjee2023orca}, with ORCA \citep{mukherjee2023orca} reaching parity performance with ChatGPT \citep{chatgpt} by distilling the reasoning traces from GPT4 \citep{openai2023gpt4}. Distinct from this line of work, \framework{} does not aim to replicate the guidance policy. Rather, our objective is to leverage generation traces derived from a guide policy to condense the search space for RL algorithms. More importantly, our goal goes beyond imitation of the guidance policy and focuses on algorithms that better optimize a reward with guidance policy feedback. \looseness=-1

\section{Preliminaries}

The sequential  nature in the task of Text generation with LLMs allows one to model it via RL. %
 In this setting, we are given a set of prompts $\{x^i\}_{i=1}^N$, and a reward function $R$ that measures some user-specified quality of the generated text. The reward $R$ can be pre-defined evaluation metrics or a learned reward model from human preference datasets. The text generation RL problem can then be defined as a token-level finite-horizon MDP $\langle \mathcal{S}, \mathcal{A}, P, R, H, \mu \rangle$ using a finite vocabulary $\mathcal{V}$. We are given a labeled dataset $\mathcal{D} = \left\{(x^i, y^i)\right\}_{i=1}^N$ of $N$ samples 
 , where $x^i$ is a prompt text and $y^i$ is the target text generation. We define $\mu \in \Delta(\mathcal{D})$ as the initial distribution over prompts, and the action space $\mathcal{A}$ as the set of tokens in our vocabulary $\mathcal{V}$.  The state space $\mathcal{S} = \cup_{h=1,\cdots,H}\mathcal{V}^h$ is the set of all possible token sequences and a state $s_h \in \mathcal{S}$ is the prompt $x$ and previously generated tokens $(a_0, a_1, \dots, a_{h-1})$, i.e.,  $s_h = (x, a_0, a_1, \dots, a_{h-1})$.
The transition function $P: \mathcal{S}\times\mathcal{A}\rightarrow \Delta(\mathcal{S})$ is a deterministic known transition function that appends the next action $a_h$ to the state $s_{h+1}$.
The time horizon $H \in \mathbb{Z}_{+}$ is the maximum generation length. 
Finally, $R: \mathcal{S} \rightarrow \mathbb{R}$ is the reward function such as the task evaluation metric or a metric learned from a preference dataset. We define our policy $\pi$ as an LLM that maps from state (i.e. prompt + partial generation) to action (next token).%

Let $d^{\pi}_{h}$ represent the state distribution of visiting a state at time $h$.
Let $d^{\pi} = \frac{1}{H} \sum_{h=0}^{H} d^{\pi}_{h}$ be the average visitation if we follow $\pi$ for $H$ steps in a trajectory.
With an LLM policy $\pi$, we define the value function and $Q$-function as $V_{h}^{\pi}(s) = \E_{\pi} [ \sum_{h'=h}^{H} {R}(s_{h'}) | s_h =s ]$ and $Q_{h}^{\pi}(s,a) = {R}(s) +  \E_{s' \sim P(\cdot|s,a)}[ V^{\pi}_{h+1}(s')]$ respectively. Finally, we define the advantage function for an LLM policy $\pi$ as $A^{\pi}(s, a) = Q^{\pi}(s, a) - V^{\pi}(s)$. 

\textbf{Guide policy $\piref$ } In our setting, we additionally assume access to a black-box LLM-based guide policy $\piref$ that can assist our policy $\pi$. The guide policy can be used to alter the initial state distribution $\mu$ and to compute the advantage function $A^{\piref}(s,a)$. In our experiments, we mainly investigate using a supervised fine-tuned (SFT) model followed by some decoding strategy (e.g., Nucleus sampling \citep{holtzman2019curious}) as $\piref$.  Note, \framework{} treats $\piref$ as a query-able, black-box model that we do not need update. This allows for $\piref$ to be any black-box model such as GPT4 or a human-expert. Our work aims to show that RLGF is capable of learning policies that are (much) better than $\piref$, and by leveraging $\piref$, it can outperform standard RL algorithm PPO.

\section{Reinforcement Learning from Guided Feedback}
\label{sec:approaches}
\begin{figure}[t]
    \centering
    \includegraphics[width=0.8\columnwidth]{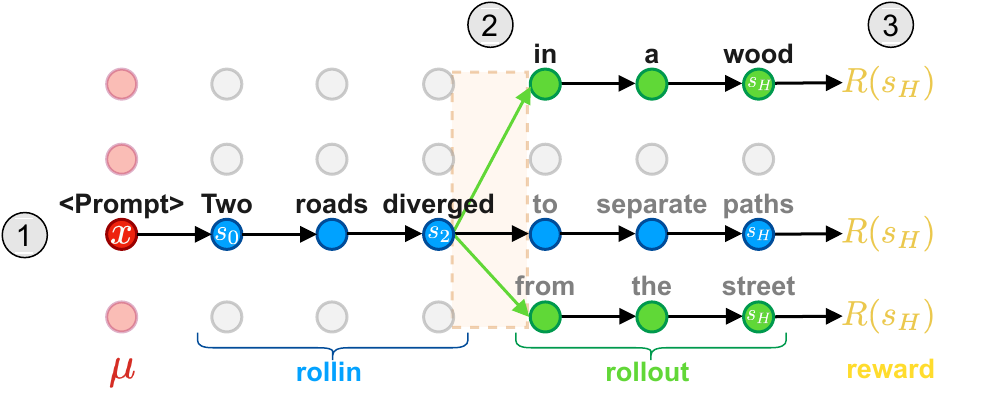}
    \caption{\framework{}'s main mechanism of incorporating guidance through interactions between two LLMs: \textcolor{Cyan}{rollin} and \textcolor{ForestGreen}{rollout} policies. (1) the \textcolor{Cyan}{rollin} policy generates a trajectory. (2) the \textcolor{ForestGreen}{rollout} policy restarts to a sampled point in the generation (i.e. $s_2$) and completes the generation. (3) the \textcolor{ForestGreen}{rollout} policy receives a score (i.e. reward) for the generation.}%
    \label{fig:rollin_rollout}
\end{figure}

Unlike other tasks studied in RL (e.g., robotics control problems), text generation  problems have two key properties: a deterministic transition function and a policy's ability to restart to any state. Because our state  is the set of previously generated tokens, we can easily alter the words in the generation (add, remove or swap), and restart our policy $\pi_{\theta}$ to any point of the generation. 

Restarts allow us to execute \textcolor{Cyan}{rollin} and \textcolor{ForestGreen}{rollout} policies as seen in \Cref{fig:rollin_rollout}. The \textcolor{Cyan}{rollin} policy is used to generate sequences that the \textcolor{ForestGreen}{rollout} policy evaluates. Specifically, we sample a prompt $x$ from our initial distribution $\mu$. We then generate an entire trajectory using our \textcolor{Cyan}{rollin} policy starting from the sampled prompt.
We combine the state-action pairs from the collected \textcolor{Cyan}{rollin} trajectory with the initial prompts -- creating a modified initial state for the \textcolor{ForestGreen}{rollout} policy.
The \textcolor{ForestGreen}{rollout} policy samples a state along the \textcolor{Cyan}{rollin} generation, restarts to this state and performs \textcolor{ForestGreen}{rollouts}. The \textcolor{ForestGreen}{rollout} policy collects a reward at the end of the generation. The \textcolor{Cyan}{rollin} and \textcolor{ForestGreen}{rollout} policies can be our LLM policy $\pi_{\theta}$, guide policy $\piref$. Depending on the choice of \textcolor{Cyan}{rollin} and \textcolor{ForestGreen}{rollout} policies, we invoke different algorithms. Note that  PPO uses $\pi_\theta$ for both rollin and rollout policies.

\paragraph{\ppo: Rollin $\pi_{\theta}$ and Rollout $\pi_{\theta}$}

Under this schematic, notice how when both the rollin and rollout policies are our current LLM policy $\pi_{\theta}$ that is being fine-tuned, the resulting RL algorithm is \ppo. That is, we would be collecting generations from a single LLM.
This configuration does not take advantage of the ability to modify the initial state distribution nor the availability of a guide policy $\piref$. \looseness=-1

\paragraph{\cpi: Rollin $\piref$ and Rollout $\pi_{\theta}$}
\begin{algorithm}[H]
    \caption{\cpi{} }
    \label{alg:finetuning_cpi}
    \vspace{-1mm}
\begin{algorithmic}[1]
    \State \textbf{Input: } $\pi_\theta$,  guide $\piref$, iterations $T$, mixing parameter $\beta\in[0,1]$, dataset $\mathcal{D} = \left\{(x^i, y^i)\right\}_{i=1}^N$
    \For{$t \in [T]$}
    \State Rollin with $(s, a) \sim \beta d^{\piref} + (1-\beta)d^{\pi^{t}_\theta}$ starting from $x\sim\mathcal{D}$ \label{line:cpi_rollin}
    \State Rollout with $\pi^t_\theta$ to collect trajectories 
    \State Update $V^{\pi^t_\theta}_\phi$ with trajectories and compute advantage estimates $A^{\pi^t_\theta}$
    \State Update $\pi_{\theta}$ using \ppo~loss with $A^{\pi^t_\theta}$
    \EndFor
    \State \Return $\pi_\theta$ 
\end{algorithmic}
\vspace{-1mm}
\end{algorithm}
The new scheme we propose is rollin with our guide policy $\piref$ and rollout with our LLM policy $\pi_\theta$. This strategy is motivated from a popular Approximate Policy Iteration algorithm \citep{bertsekas2011approximate}: Conservative Policy Iteration (CPI) \citep{kakade2002cpi}. CPI proposes to use a diverse initial state distribution to address the exploration issue in PG methods. Particularly, it proposes to use an initial state distribution that covers some high-quality policy's state distribution.
The first key idea of \cpi{} is to take advantage of a guide policy $\piref$ to provide an enlarged initial state distribution -- so that the rollout policy, $\pi_\theta$, can visit diverse and relevant states it would otherwise not visit. The second key idea of \cpi{} is using a mixture policy with state distribution $\beta d^{\piref} + (1-\beta)d^{\pi_\theta}$, for rollin (see \cref{alg:finetuning_cpi} Line~\ref{line:cpi_rollin}). This ensures that with probability $(1-\beta)$, \cpi{} is executing the default PPO update, making sure \cpi{} maintains the benefits of PPO and never underperforms PPO. \looseness=-1

\paragraph{\agg: Rollin $\pi_\theta$ and Rollout $\piref$}

\begin{algorithm}[H]
    \caption{\agg{}}
    \label{alg:finetuning_agg}
\begin{algorithmic}[1]
    \State \textbf{Input: } $\pi_\theta$,  guide $\piref$, iterations $T$, mixing parameter $\beta\in[0,1]$, dataset $\mathcal{D} = \left\{(x^i, y^i)\right\}_{i=1}^N$
    \For{$t \in [T]$}
    \State Rollin with $(s,a)\sim(1-\beta)d^{\pi^{t}_\theta} + \beta d^{\piref}$ starting from $x\sim\mathcal{D}$
    \State Rollout with $\piref$ to collect trajectories 
    \State Update $V^{\piref}_\phi$ with trajectories and compute advantage estimates $A^{\piref}$ 
    \State Update $\pi_\theta$ using \ppo~ loss with $A^{\piref}$
    \EndFor
    \State \Return $\pi_\theta$
\end{algorithmic}
\end{algorithm}

The next scheme performs rollin with our LLM policy $\pi_\theta$  and rollout with our guide policy $\piref$ -- the opposite of \cpi. This scheme is an interactive imitation learning algorithm, \agg{} \citep{sun2017aggrevated}, a differentiable policy gradient version of AggreVaTe (Aggregate Values to Imitate \citep{ross2014aggrevate}) as seen in \cref{alg:finetuning_agg}. \agg~ is an API algorithm similar to CPI and also uses a mixture policy with state distribution $\beta d^{\piref} + (1-\beta) d^{\pi_{\theta}}$ for rollin. This algorithm first generates rollins with the mixture policy to collect sequences. 
Then \agg~generates rollouts with the guide policy and evaluates the quality of the generated rollouts. It then uses the rollouts to train a value network $V^{\piref}_\phi$ that measures the reward-to-go of $\piref$, which in turn is used to construct the advantage of $\piref$: $A^{\piref}$. With this advantage  $A^{\piref}$, \agg{} updates the policy like PPO (i.e., update $\pi_\theta$ so that it increases the probabilities of selecting actions with larger $A^{\piref}$ ). Intuitively, the algorithm aims to learn the policy $\argmax_{a} A^{\piref}(s,a)$,
which ensures that that the LLM policy $\pi_\theta$ can be \textit{at least} as good as or better than the guide policy $\piref$.\looseness=-1

\paragraph{\lolsd: combines \cpi~and \agg}

\begin{algorithm}[h]
    \caption{\lolsd}
    \label{alg:finetuning_hard_lols}
\begin{algorithmic}[1]
    \State \textbf{Input: } $\pi_\theta$,  guide $\piref$, iterations $T$, dataset $\mathcal{D} = \left\{(x^i, y^i)\right\}_{i=1}^N$
    \State Run $\pi_\theta^1 = \agg{}(\pi_\theta, \piref, \alpha T, \beta_1, \mathcal{D})$
    \State Run $\pi_\theta^2 = $ \cpi{}$(\pi_\theta^1, \piref, (1-\alpha)T, \beta_2, \mathcal{D})$
    \State \Return $\pi_\theta^2$
\end{algorithmic}
\end{algorithm}

Given the previous approaches of interaction, we can come up with multiple ways to combine \ppo, \cpi, and \agg. In \cref{alg:finetuning_hard_lols}, we present Direct and Differentiable Locally Optimal Learning to Search (\lolsd), which is a simple approach to combine the previous methods. \lolsd{} is a differentiable policy gradient version of Locally Optimal Learning to Search (\lol)\citep{chang2015learning} and addresses limitations of how \lol{} combines \ppo, \cpi, and \agg. The original formulation of \lol{} requires computing cost-sensitive classification similar to AggreVaTe; instead we take inspiration from \agg{}'s differentiable approach to develop a differentiable version of \lol. Furthermore, \lol{} (\cref{alg:finetuning_lols}) has a mixing probability parameter $\alpha$ which directly merges the advantage function between \ppo{} and \agg{}, leading to theoretical issues. \lolsd{} removes this mixing probability and replaces it with a mixing time variable $\alpha$ that decides how many iterations to perform \agg{} before switching to \cpi{}. This simple modification not only makes \lolsd{} more practical to optimizing LLMs, but also fixes \lol's issue arising from interweaving guidance. Thus \lolsd{} should be understood as a more practical and more principled alternative of \lol. \looseness=-1

\section{Theoretical Justification}%
\label{sec:theory}
In this section, we provide theoretical justification for various rollin and rollout schemes mentioned in \cref{sec:approaches}.
Each algorithmic scheme takes advantage of a guide policy $\piref$, the ability to restart the policy to any state, and access to the reward signal. Our theoretical justification are derived from the original algorithms that each method has built upon.

\paragraph{Interactive Imitation Learning: \agg}
In our interactive IL setting, we assume access to the ground truth reward and to a guide policy $\piref$ that may not necessarily be an expert policy $\pi^\star$ (i.e. optimal at the task). 
Our \agg~ (\cref{alg:finetuning_agg}) implementation is a modification of the original \agg~\citep{sun2017aggrevated} to incorporate a \ppo~policy gradient loss. The overall idea is to perform policy gradient updates on the loss function $\ell_t(\pi):= \mathbb{E}_{s \sim d^{\pi^t} }  \mathbb{E}_{a\sim \pi(\cdot | s )}[A^{\piref}(s, a)]$, where $\pi^t$ is our latest learned policy. 
We can define the average-regret and best policy performance in our policy class over $T$-iterations as: 
\begin{align*}
    \epsilon_{\mathrm{regret}} = \frac{1}{T}\left(- \sum_{t=0}^T \ell_t(\pi^t) + \max_{\pi\in \Pi} \sum_{t=0}^T \ell_t(\pi) \right)
    &&
    \epsilon_{\mathrm{class}} = \max_{\pi \in \Pi} \frac{1}{T} \sum_{t=0}^T \mathbb{E}_{s\sim d^{\pi^t}} \left[ A^{\piref}(s, \pi(s)) \right]. 
\end{align*} 
If the gradient update procedure achieves no-regret, i.e., $\epsilon_{\mathrm{regret}} \to 0$ as $T\to \infty$, \agg{} achieves the following guarantee; there exists $t\in [T]$, such that:  %
\begin{align*}
 V^{\pi^t} \geq V^{\piref} + H \epsilon_{\mathrm{class}}.
\end{align*} 
When the guide policy is included in our policy class $\piref \in \Pi$, e.g., when our policy $\pi_\theta$ and our guide $\piref$ have the same GPT2 model architecture, then our $\epsilon_{\mathrm{class}}$ term is guaranteed to be non-negative. Furthermore, this term is positive when $\piref$ is not globally optimal with respect to its advantage function (i.e., $\max_{a} A^{\piref}(s,a)$ can be positive). Thus when $\epsilon_{\mathrm{regret}}\to 0$ (i.e., no-regret), \agg{} guarantees to learn a policy $\pi_t$ that outperforms the guide policy by a margin. This was originally confirmed empirically in \cite{sun2017aggrevated} and is also confirmed in our experiments. With our SFT model with nucleus sampling as $\piref$, \agg{} learns a policy $\pi^t$ outperforming $\piref$. \looseness=-1%

\paragraph{Reinforcement Learning with better restart distribution: \cpi}

Although \agg{} is capable of outperforming $\piref$, it is an imitation learning algorithm, meaning by design, its performance is limited by the performance of $\piref$. In contrast, RL has the potential to learn the near optimal policy, but popular RL approaches suffer from a lack of exploration. We propose to leverage rollin's with the guide policy to overcome RL's exploration issues. \cpi~\cref{alg:finetuning_cpi} implements this idea using a \ppo~loss. We can interpret the rollin policy distribution with the guide policy, as a restart distribution that alters the initial distribution of our policy, i.e., $\mu_{\mathrm{mix}} := (1-\beta) \mu + \beta  d^{\piref}$, where recall $\mu \in \Delta(\mathcal{D})$ is the original initial state distribution over our data.

Policy gradient theory \citep{kakade2002cpi, bagnell2003policy, agarwal2019reinforcement, agarwal2021theory} ensures that
as long as a near optimal policy is covered by the restart distribution, we can learn to perform as well as the near optimal policy. More formally, consider the special case where $\beta = 1/2$, and $\pi^\star$ is the globally optimal policy; and assume that at some iteration $t$ one-step local improvement over $\pi^t$ is small, i.e., $\mathbb{E}_{s,a\sim d_{\mu_{\mathrm{mix}}}^{\pi^t} }\left[ \max_{a} A^{\pi^t}(s,a)\right] \leq \epsilon$, then with some small $\epsilon$ we have: 
\begin{align*}
&V^{\pi^t} \geq V^{\pi^\star} - O\left(H^2  \max_{s}\left( \frac{d^{\pi^\star}(s)}{ d^{\piref}(s) } \right) \epsilon \right)
\end{align*}
We refer readers to the proof of theorem 6.2 in \cite{kakade2002cpi}.
Note that compared to the result from \agg, we are able to compare against the globally optimal policy $\pi^\star$ under %
the condition that $\piref$'s state distribution covers $\pi^\star$'s state distribution %
(i.e., the guide policy has a good sense of what states $\pi^\star$ will likely visit). In our experiments, we mainly use a SFT model with nucleus sampling as our guide policy $\piref$. While we do not expect the SFT policy $\piref$ is as good as the optimal $\pi^\star$, it is reasonable to expect that $d^{\piref}$ provides coverage to $d^{\pi^\star}$. Our experiments verify that restarting based on states from $d^{\piref}$ improves the performance of PPO.

\paragraph{Combine Reinforcement Learning and Imitation Learning: \lolsd}
\label{sec:lols}
\lolsd~is the simplest approach to combine \agg{} and \cpi{}. This algorithm runs \agg{} for a fixed period of time and then \cpi{} for the remaining time. If our policy gradient algorithm is Trust-region policy optimization (TRPO) \footnote{in our experiments, instead of using TRPO, we use PPO -- a scalable version of TRPO that is more suitable for high-dimensional problems. However we emphasize the TRPO and PPO use the same principle for policy optimization: make conservative policy update \citep{kakade2002cpi} to ensure monotonic improvement.} \citep{schulman2015trust} or CPI \citep{kakade2002cpi}, then our algorithm has a guaranteed monotonic policy improvement. This means that upon convergence, we achieve two properties: (1) our learned policy is at least as good or better than the guide policy $\piref$, (2) our policy is locally optimal, i.e., the local one-step improvement, $\mathbb{E}_{s,a\sim d^{\pi}_{\mu_{mix}}} \left[ \max_{a} A^{\pi}(s,a)\right]$, has to be small (otherwise TRPO and CPI can keep improving).

There exist several algorithms in the literature that combine RL and IL \citep{cheng2018fast, sun2018truncated, chang2015learning, rajeswaran2017learning, nair2018overcoming}. 
The key difference between \lolsd{} and \lol{} is how \cpi{} and \agg{} is combined. \lol{} uses a mixing probability $\alpha$ to combine our $\pi_\theta$ and the guide policy $\piref$ advantage function $\alpha A^{\pi^t_\theta} + (1 - \alpha) A^{\piref}(s,a)$; whereas \lolsd{} uses a mixing time parameter $\alpha$ to decide when to switch from doing \agg{} to \cpi{} for the remainder of training. 
\lol{} can achieve the property of outperforming better than $\piref$ and also being locally optimal, but \emph{only under} the assumption that the following gap is small: 
\begin{align*}
    \forall \pi: & \Big\lvert \mathbb{E}_{s\sim d^\pi} \left[\max_{a} A^{\piref}(s,a) + \max_{a} A^{\pi}(s,a)\right]  -  \mathbb{E}_{s\sim d^\pi} \max_{a}\left[ A^{\piref}(s,a) + A^{\pi}(s,a)\right]  \Big\rvert \leq \varepsilon,
\end{align*} with some small $\varepsilon$. However, such a gap can exist in practice and does not vanish even with enough training data. Intuitively this gap is non-trivial when the one-step improvement over $\pi$ contradicts with the one-step improvement over $\piref$.  
The simplest approach \lolsd{} works the best, and achieves the guarantee that \lol{} aimed for without the additional assumption of the above gap being small. \looseness=-1

\section{Experiments}

\begin{table}[tb]
    \centering
    \resizebox{\textwidth}{!}{
    \begin{tabular}{@{}lccc||ccc@{}}
        \toprule
        & \multicolumn{3}{c}{\textbf{IMDB Sentiment}} & \multicolumn{3}{c}{\textbf{CommonGen}} \\
        \textbf{Algorithms} & \multicolumn{3}{c}{\textit{Semantic and Fluency Metrics}} & \multicolumn{3}{c}{\textit{Lexical and Semantic Metrics}} \\
        \cmidrule(lr){2-4}\cmidrule(lr){5-7}
        & \textbf{Sentiment Score}  & Perplexity  & Output-Perplexity  &Bleu-4  & \textbf{CIDEr-D} & \textbf{SPICE} \\
        & ($\uparrow$) & ($\downarrow$) & ($\downarrow$) & ($\uparrow$) & ($\uparrow$) & ($\uparrow$) \\
        \cmidrule(lr){1-1}\cmidrule(lr){2-4}\cmidrule(lr){5-7}
        Zero-Shot & 0.48 $\pm$ 0.00 & 32.55 $\pm$ 0.00 & 5.64 $\pm$ 0.00 & 0.00 $\pm$ 0.00 & 6.02 $\pm$ 0.55 & 15.02 $\pm$ 0.40 \\
        \bc{}     & 0.55 $\pm$ 0.00 & 35.67 $\pm$ 0.00 & 6.19 $\pm$ 0.00 & 22.31 $\pm$ 0.12 & 14.32 $\pm$ 0.15 & 31.73 $\pm$ 0.34\\
        \cmidrule(lr){1-1}\cmidrule(lr){2-4}\cmidrule(lr){5-7}
        \bcppo{}    & \textbf{0.97} $\pm$ \textbf{0.01} & 44.92 $\pm$ 1.78  & 3.17 $\pm$ 0.62 & 27.98 $\pm$ 0.32 & 16.91 $\pm$ 0.29 & 32.61 $\pm$ 0.06          \\
        \bccpi{}    & \textbf{0.97} $\pm$ \textbf{0.01} & 44.83 $\pm$ 2.10  & 3.34 $\pm$ 0.80 & 28.48 $\pm$ 0.24 & 16.94 $\pm$ 0.53 & 32.75 $\pm$ 0.21 \\
        \bcagg{}    & 0.95 $\pm$ 0.03 & 52.56 $\pm$ 5.38  & 5.04 $\pm$ 2.30 & 28.14 $\pm$ 0.31 & 16.90 $\pm$ 0.09 & 32.44 $\pm$ 0.02 \\
        \bclol{}    & 0.93 $\pm$ 0.05 & 53.30 $\pm$ 16.70 & 3.44 $\pm$ 4.96 & 28.15 $\pm$ 0.16 & 16.91 $\pm$ 0.22 & 32.80 $\pm$ 0.20 \\
        \bclolsd{}  & \textbf{0.97} $\pm$ \textbf{0.00} & \textbf{43.88} $\pm$ \textbf{2.37}  & \textbf{2.92} $\pm$ \textbf{0.13} & \textbf{28.54} $\pm$ \textbf{0.12} & \textbf{16.96} $\pm$ \textbf{0.18} & \textbf{32.83} $\pm$ \textbf{0.09} \\
        \bottomrule 
    \end{tabular}}
    \vspace{1mm}
    \caption{\textbf{IMDB and CommonGen Results:} We compute the mean and standard deviation over 3 seeds for both the IMDB and the CommonGen tasks. For our reward function each task we use the bold metric(s). The zero-shot model is the performance of the pretrained model used for IMDB and CommonGen,  GPT-2 and T5 respectively. \texttt{SFT}+\texttt{Alg} indicates running \texttt{Alg} after supervised finetuning. \texttt{SFT+nucleus} is used as our guide policy $\piref$ for all experiments.}
    \label{tbl:main_results}
    \vspace{-3mm}
\end{table}

We perform all of our experiments using a modified \ppo{} objective $J_{ppo}$ \citep{ouyang2022training, wu2016google}. This objective combines the original $\ppo{}$ objective with a maximum-likelihood estimation (MLE) objective of the ground-truth dataset's $\mathcal{D}$ references: %

\begin{align*}
J_{ppo}(\pi_\theta) = \mathbb{E}_{(s,a) \sim \pi_\theta} \Big[ R(s) - \lambda \text{KL}(\pi_\theta(a|s) || \pi_0(a|s)) \Big] + \eta \mathbb{E}_{(s,a) \sim \mathcal{D}} \Big[ \log\pi_\theta(a|s)\Big],
\end{align*}

where $\lambda$ is the $\text{KL}$ coefficient and $\eta$ is the MLE coefficient. For all of our proposed \framework{} algorithms discussed in section \ref{sec:approaches} we consider setting  $\piref$ to the supervised fine-tuned model ($\bc{}$) with nucleus sampling for decoding (i.e., $\piref =$\texttt{SFT+nucleus}). We treat \texttt{SFT+nucleus} as a black-box model that we can only query for text generation and do not perform updates to  it. By using \texttt{SFT+nucleus} as our guide policy, we run all of our experiments under the exact same conditions as those of RLHF. Note, RLHF already requires keeping \bc{} to compute the KL constraint, $KL(\pi_\theta||{\color{blue}{\pi_0}})$, in $J_{ppo}$.

\paragraph{Task Details} 
In our experiments, \textit{perplexity} measures how likely our learned model, $\pi_\theta$, is to generate the references in the task dataset, whereas \textit{output perplexity} computes how likely a general LLM (e.g. GPTJ) is to generate the generations from our learned policy, $\pi_\theta$. Both perplexity metrics have been reported as a measure of fluency \citep{fedus2018maskgan, ramamurthy2022rl4lm}.

We perform experiments on three tasks. IMDB is the first task and the objective is to generate fluent and positively sentiment-ed text continuations for IMDB \citep{maas2011learning} movie reviews prompts. We use a sentiment classifier \citep{sanh2019distilbert} as our reward function that is trained on review texts and sentiment labels from the dataset, which then provides sentiment scores indicating how positive a given piece of text is. For training supervised \bc{} baselines, we consider only the examples with positive labels. We chose \gpt{} \citep{radford2019language} as the base language model (LM) for this task. We evaluate all algorithms on three metrics: sentiment reward score, perplexity, and output-perplexity. 

Next, we consider CommonGen \citep{lin-etal-2020-commongen}, a challenging constrained, text generation task that tests the ability of generative common sense reasoning. We optimize the SPIDER \citep{liu2017improved} reward function, a weighted combination of the CIDEr-D and SPICE metric. We chose \texttt{T5-base} \citep{2020t5} as our base LLM and prefixed each concept set input with: "generate a sentence with:". We report four  metrics: BLEU \citep{papineni2002bleu}, CIDEr-D \citep{vedantam2015cider}, and SPICE \citep{anderson2016spice}. For IMDB and CommonGen, we perform one epoch of supervised finetuning for our \bc{} models.%

The final task we consider is Reddit TL;DR summarization dataset \citep{volske-etal-2017-tl} where the objective is to generated summaries. We use the filtered dataset with additional human preference data used in \cite{stiennon2020learning}. The base LLM that we use for this task is GPT-J \citep{wang2021gpt} and we train all models in our algorithms using LoRA adapters \citep{hu2021lora}. We evaluate all algorithms on 5 metrics: reward score, perplexity, output-perplexity, win rate and Rouge \citep{lin2004rouge}. For win rate, we use the open source Llama2-13B-chat \citep{touvron2023llama} model as our evaluator model. We compare all algorithm generations to the preferred summary references. For our \bc{} model, we use an open-source GPT-J model\footnote{\url{https://huggingface.co/CarperAI/openai_summarize_tldr_sft}}.Refer to \cref{appendix:task}, for the exact Win Rate prompt, example evaluations and implementation details.

\begin{table}[t]
    \centering
    \resizebox{\textwidth}{!}{
    \begin{tabular}{@{}lccccccccc@{}}
        \toprule
        & \multicolumn{7}{c}{\textbf{TL;DR Summarization}} \\
        \textbf{Algorithms} & \multicolumn{7}{c}{\textit{Semantic and Fluency Metrics}} \\
        \cmidrule(lr){2-8}
        & \textbf{RM Score}  & Perplexity  & Output-Perplexity & Win Rate & Rouge 1 & Rouge 2 & RougeL \\
        & ($\uparrow$) & ($\downarrow$) & ($\downarrow$) & ($\uparrow$) & ($\uparrow$) & ($\uparrow$) & ($\uparrow$)  \\
        \cmidrule(lr){1-1}\cmidrule(lr){2-8}
        Zero-Shot & 1.57 & 14.07  & 11.51 & 44.12\% & 0.27 & 0.07 & 0.18 \\ 
        \bc{}     & 5.68 & 14.09 & 12.81 & 44.29\% & 0.34  & 0.25 & 0.25 \\
        \bestofn{}\,($N=8$) & 5.98 & 14.09 & 12.86 & 47.60\% & 0.36  & 0.13  & 0.27 \\
        \cmidrule(lr){1-1}\cmidrule(lr){2-8}
        \bcppo{}  & 6.01 & 15.05 & 17.67 & 54.25\% & 0.35 & 0.13 & 0.27 \\
        \bccpi{}  & 6.11 & 14.53 & 16.15 & 55.01\% & 0.36 & 0.14 & 0.27  \\
        \bcagg{}  & 5.93 & 14.69 & 16.41 & 48.98\% & 0.36 & 0.15 & 0.29 \\
        \cmidrule(lr){1-1}\cmidrule(lr){2-8}
        \bcppo{}\,($N=8$)  & 6.20 & 14.87 & 16.53 & 57.53\% & 0.36 & 0.15 & 0.27 \\
        \bccpi{}\,($N=8$)  & \textbf{6.52} & \textbf{13.42} & \textbf{15.23} & \textbf{60.30}\% & \textbf{0.38} & 0.15 & \textbf{0.28}  \\
        \bcagg{}\,($N=8$)  & 6.11 & 13.53 & 15.61 & 54.12\% & 0.37 & \textbf{0.16} & \textbf{0.28} \\
        \bottomrule 
    \end{tabular}}
    \vspace{1mm}
    \caption{\textbf{ TL;DR Summarization Results:} We report the mean over 1 seed. Our RM Score is under our trained preference reward model and the Win Rate is evaluated by Llama2-13B-Chat. We use \texttt{SFT+nucleus} as $\piref$. We also report Best-of-8 results with our trained policies.}
    \label{tbl:tldr_results}
    \vspace{-2mm}
\end{table}

\subsection{Experimental Results}
\paragraph{\framework{} vs. RLHF Performance} \cref{tbl:main_results} and \cref{tbl:tldr_results} compares all of the \framework{} algorithms proposed in \cref{sec:approaches} against standard RLHF algorithms and baselines. For all tasks, our $\piref$ is \texttt{SFT+nucleus} which is sub-optimal, performing worse than all RL based algorithms across most lexical and semantic metrics. 
Utilizing this $\piref$, for IMDB, \lolsd{} outperforms \ppo{} on all metrics while \cpi{} outperforms \ppo{} on both semantic reward and perplexity, and for CommonGen, \lolsd~ outperforms \ppo~ in all metrics including the ones that are not included in the reward function. Finally, for TL;DR summarization we see that \cpi{} performs better than \ppo{} as well as a competitive baseline, \bestofn{} \citep{dubois2023alpacafarm}. Furthermore, when applying \bestofn{} inference on our trained policies, we see that \cpi{} improves even more beyond \ppo{}. Notably, with or without best-of-N procedure, \cpi{} outperforms \ppo{} on all metrics.

Supporting our justification from \cref{sec:theory}, \agg{} improves beyond our guide policy, providing an alternative as a warm-starting methodology to warm-starting with \bc{}. \cpi{}, on the other hand, is better than or competitive to our RL baseline demonstrating a simple, yet powerful alternative to \ppo{} as the RL procedure. Even in practice, we observe the benefit of restarting from an initial state distribution that better covers an optimal policy's state distribution. The combination of these two, \lolsd{}, achieves the best of both worlds and fully leverages the capabilities of utilizing a guide policy.

\begin{figure}[b]
    \centering
    \includegraphics[width=0.65\textwidth]{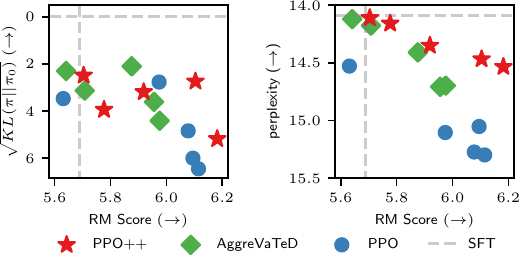}
    \caption{ We investigate the reward optimization, kl-constriant, and fluency trade-off in our TL;DR summarization task. The dashed line represents our $\bc$ policy's performance across each metric. Both \cpi{} and \agg{} learn a policy that has a better trade-off than \ppo{}.}
    \label{fig:tradeoff}
\end{figure}
\paragraph{Reward Optimization Tradeoff}
In \cref{fig:tradeoff} we evaluate how well \framework{} algorithms trade-off optimizing the reward while minimizing the perplexity and kl-constraint $\sqrt{\text{KL}}$. For fair comparisons, we kept $\lambda$ and $\eta$ the same across all algorithms. For both plots, the top right corner indicates the policy has both high reward and low perplexity and low divergence from $\pi_0$. For each algorithm we plot 5 checkpoints ranging from 20 to 100 iterations.\cpi{} mostly matches or has higher reward than \ppo{} while maintaining a lower perplexity. Separately, \agg{} trade-offs reward for perplexity, and has comparable reward scores as \ppo{} while drastically reducing its perplexity. For the kl-constraints plot on the left of \cref{fig:tradeoff} we see that although \ppo{} has a set of points with high reward, most of these points also have high KL divergences. Whereas, a subset of \cpi{} matches or has higher reward than \ppo{} while having a lower kl-constraint. \looseness=-1

\begin{figure}[t]
    \centering
    \includegraphics[width=0.8\textwidth]{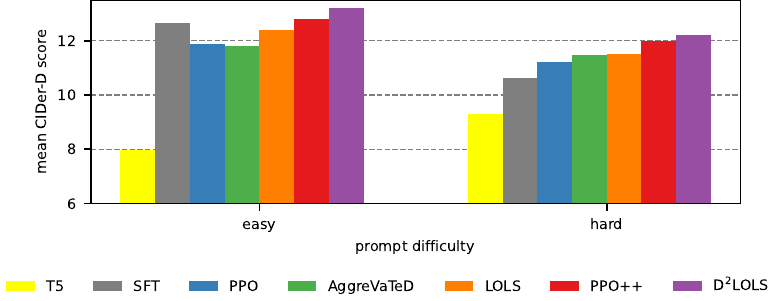}
    \caption{Comparison of CIDer-D scores grouped by prompt difficulty on CommonGen. The performance gap between easy and hard prompts is evident for \bc, and \cpi, while our proposed algorithms \agg{}, \lol{} and \lolsd{} exhibit a significantly smaller gap, showcasing their effectiveness on challenging prompts.} %
    \label{fig:scores_by_difficulty.pdf}
    \vspace{-3mm}
\end{figure}
\begin{figure}[t]
    \centering
    \includegraphics[width=1.\textwidth]{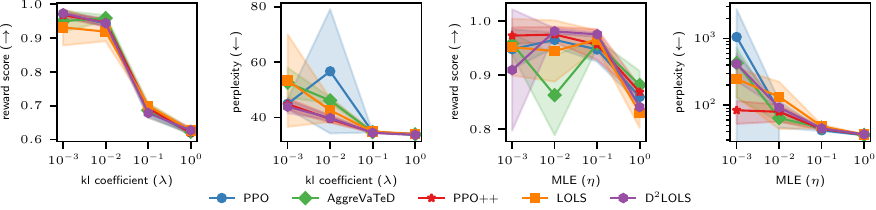}
    \vspace{-2mm}
    \caption{$J_{ppo}$ KL coefficient ($\lambda$) and MLE coefficient ($\eta$) ablation. We show the sensitivity of PPO and RLGF algorithms to each regularization term in the objective. Note that all RL algorithms are robust to changes in KL coefficient with relatively minor changes in the Perplexity while being more sensitive to changes in MLE objective (Right) with blowups in the perplexity.}
    \label{fig:kl_ablation.pdf}
    \vspace{-2mm}
\end{figure}
\paragraph{\framework{} Performance on Difficult Prompts}
Our evaluation was carried out on the CommonGen task where we categorized the prompts based on their difficulty level.
For CommonGen, we classify the prompts into \textit{easy} and \textit{hard} based on the number of unseen concepts in the prompt. Specifically, we categorized prompts with 3 concepts as easy and more than 3 concepts as hard. \cref{fig:scores_by_difficulty.pdf} presents a comparison of scores for different algorithms grouped by prompt difficulty. The results reveal a notable performance gap between easy and hard prompts for algorithms such as \bc{} and \ppo{}, whereas our proposed algorithms \cpi{}, \agg{}, \lol{} and \lolsd{} exhibit a smaller gap, with \lolsd{} having the least gap %
. In other words, even on challenging prompts, our interactive algorithms produce better text continuations. See \cref{appendix:commongen} for example generations.

\paragraph{MLE and KL coefficient Sensitivity} We test the sensitivity of PPO and RLGF algorithms to two regularization hyperparameters in the $J_{ppo}$ objective, namely the KL coefficient, $\lambda$, and the MLE coefficient, $\eta$. The left 2 plots in \cref{fig:kl_ablation.pdf} show the reward and perplexity when we keep $\eta$ fixed and vary $\lambda$ while the right 2 show the performance when we keep $\lambda$ fixed and vary $\eta$. As shown in the left two figures, all RL algorithms are robust to varying KL coefficients. We see that when we varying $\lambda$, while our algorithms \cpi{}, \lolsd{} and the baseline \ppo{} has similar rewards, our algorithms consistently maintain a lower (or equal) perplexity than \ppo.  From the right two figures, we observe much more instability on perplexity when relaxing our MLE regularization with both PPO and RLGF algorithms' perplexities blowing up. Note that when increasing $\eta$, our algorithm \cpi{} consistently has higher rewards and lower (or equal) perplexity than \ppo. %

\section{Conclusion and Future Work}
We presented a unifying framework of incorporating a guide policy to enhance reinforcement learning for natural language generation. Through theoretical justification and experimental validation, we demonstrate that our \framework{} framework can outperform PPO for fine-tuning LLMs. Our proposed algorithms \cpi{} and \lolsd{} only require black-box access to the guide policy and are conceptually simple and easy to implement based on \ppo.
While in our experiment, we demonstrate that supervised fine-tuned models with standard decoding strategies is a good candidate of the guide policy, our framework is general enough to leverage any large LLMs as the guide policy, including those that are not open-sourced. 
Finally, \framework{}'s contributions to the broader large language model literature is complementary to model enhancements, dataset improvements, and prompting discoveries such as in-context prompting. We leave it to exciting future work to test the full capabilities of bootstrapping the state-of-the-art advancements in each research direction with \framework{} to improve reinforcement learning for natural language generation.\looseness=-1

\section{Acknowledgements}
We would like to acknowledge the support of NSF under grant IIS-2154711. Jonathan Chang is supported by LinkedIn under the LinkedIn-Cornell Grant. Kiante Brantley is supported by NSF under grant No. 2127309 to the Computing Research Association for the CIFellows Project. Rajkumar Ramamurthy is funded by the Federal Ministry of Education and Research of Germany and the state of North-Rhine Westphalia as part of the Lamarr-Institute for Machine Learning and Artificial Intelligence.

\bibliographystyle{iclr2024_conference}

\newpage
\appendix
\onecolumn

\clearpage
\newpage

\section{Additional Related Work}
\label{appendix:related_work_full}

\textbf{LLM Alignment } Using RLHF is one idea of aligning LLM with human preferences. The RLHF objective incorporates a KL constraint and is equivalent to minimizing the reverse KL between KL-control distribution and the learner. Minimizing some divergence between policy used for the KL-control and learner policy has been proposed for LLM alignment.
\citep{korbak2022reinforcement, khalifa2020distributional, go2023aligning} propose alignment ideas the attempt to minimize various divergence inspired from maximize entropy RL \citep{haarnoja2017reinforcement, haarnoja2018soft} and Distributional Policy Gradient (DPG) \citep{barth2018distributed}. Depending on the chosen divergence, the desired policy behavior may be easy or hard to obtain.  Another collection ideas for alignment focus on aspects of the supervised learning data, for example currating the collected data \citep{zhou2023lima, chung2022scaling}.

\textbf{Restart Distribution } On-policy RL algorithms are not able to take advantage of past visited states. But incorporating the ability to reset to any arbitrary state allows on-policy methods to create new states from past visited states \citep{tavakoli2018exploring}. The core of the idea is to use past visited states to modify the initial state distribution. Our work introduces \cpi~ which is an algorithm that has no prior over past visited states but \citep{tavakoli2018exploring} considers incorporating priories to help decide how to prioritize past visited states to incorporate into the initial state distribution. \citep{agarwal2020optimality} showed theoretically that the initial state distribution helps with exploration. Modifying the initial state distribution using restart has seen success in Montezuma Revenge Atari 2600  (a hard exploration problem) and Atari 2600 games more broadly\citep{popov2017data, salimans2018learning, ecoffet2019go, florensa2017reverse}.

\textbf{NLP with Human Feedback}
Learning from human feedback has been studied in the past in the  context of bandit feedback \citep{nguyen2017reinforcement, sokolov2016bandit}, pairwise feedback \citep{scheurer2023training, chen2023improving} and other feedback forms \citep{kreutzer2018can, sumers2021learning, hancock2018training, wu2021recursively}.
RLHF from has been an active area of research employing RL as the main strategy to align LMs with human preferences \citep{ouyang2022training, bai2022training, bakker2022fine, chatgpt, nakano2021webgpt, wu2021recursively, stiennon2020learning, ziegler2019fine}. A remarkable result in this line of work is ChatGPT~\citep{chatgpt}. The general process involves learning a preference reward model induced by human preferences and then finetuning with RL using this learned preference model. 

\textbf{LLM Finetuning from AI Feedback: }
Despite being easier to collect than expert data, high-quality human preference data collection is a key bottleneck of scaling RL finetuning for LLMs. A growing body of work enlists the help of LLMs to augment various parts of the RLHF procedure. ConstitutionalAI and RLAIF \citep{bai2022constitutional, lee2023rlaif} explores using LLMs to generate preference datasets to do reward model training on while \citep{roit2023factually, yang2023rlcd, kwon2023reward} finds directly generating reward signals from another LLM to be effective. Separate from this literature, we investigate utilizing direct LLM feedback during the generation process, reminiscent of RL algorithms utilizing expert interactive feedback.

\textbf{RL for Text Understanding and Generation:} RL has been used to train text generation models for dialogue~\citep{li2016deeprldialogue}, 
text simplification~\citep{zhang-lapata-2017-sentence}, 
machine translation~\citep{kiegeland2021revisiting, wu2016google, shen2015minimum}, image captioning~\citep{Ren_2017_CVPR}, question generation~\citep{pang2021text}. RL has also been used to create models that take actions given a text such as for instruction following~\citep{hermann2017grounded,misra-etal-2017-mapping}, text games~\citep{narasimhan-etal-2015-language, cote2019textworld, ammanabrolu2018playing}, and code generation~\citep{zhong2017seq2sql}. These methods typically use policy gradient based RL. Recently, \citep{ramamurthy2022rl4lm} studied online RL for text generation across a wide range of tasks, specifically studying Proximal Policy Optimization (PPO) \citep{schulman2017proximal}. Although the results comparing RL and SL are mixed, we build upon their work and show the benefit of RL and ultimately \framework{} outperforming SL and RL. Separately, \citep{snell2022offline} studies offline RL in the context of text generation whereas our work studies the online case.

\clearpage
\newpage

\section{Additional Algorithms}
A detailed algorithm for \lol{} showing how to combine reinforcement learning and imitation learning differently than \lolsd{}. Rather than setting $\alpha$ to be the stopping time to switch from \agg{} to \cpi{}, we have a mixing probability of combining \agg{} and \cpi{} at every iteration, $\alpha A^{\pi^t_\theta} + (1 - \alpha) A^{\piref}(s,a)$. As discussed in \Cref{sec:lols}, we find that \lol{} underperforms \lolsd{}, even in practice.
\begin{algorithm}[h]
    \caption{\lol: combine \ppo~and \agg}
    \label{alg:finetuning_lols}
\begin{algorithmic}[1]
    \State \textbf{Input: } $\pi_\theta$,  reference $\piref$, iterations T, dataset $\mathcal{D} = \left\{(x^i, y^i)\right\}_{i=1}^N$
    \State \textbf{Input: } mixing parameter $\beta_{1}\in[0,1]$, mixing parameter $\beta_{2}\in[0,1]$, mixing prob $\alpha$

    \For{t = 0,1,\ldots,T-1}
    \State \LeftComment{\cpi}
    \State Rollin with $\beta_1\piref + (1-\beta_1)\pi^t_\theta$ starting from $x\sim\mathcal{D}$
    \State Rollout with $\pi^t_\theta$ to collect trajectories
    \State Update $V^{\pi^t_\theta}_\phi$ with trajectories and compute advantage estimates $A^{\pi^t_\theta}$
    
    \State \LeftComment{\agg} 
    \State Rollin with $\beta_2\pi^t_\theta + (1-\beta_2)\piref$ starting from $x\sim\mathcal{D}$
    \State Rollout with $\piref$ to collect trajectories
    \State Update $V^{\piref}_\phi$ with trajectories and compute advantage estimates $A^{\piref}(s,a)$ 
    
    \State \LeftComment{Mix Update} 
    \State Update $\pi_\theta$ using \ppo~ loss with $\alpha A^{\pi^t_\theta} + (1 - \alpha) A^{\piref}(s,a)$
    \EndFor
\end{algorithmic}
\end{algorithm}

\clearpage
\newpage
\section{Additional Experimental Details}
\subsection{KL Reward Constraint}
In addition to sequence-level task rewards, per-token KL rewards are applied to prevent the policy $\pi$ from deviating too far from the pre-trained LM $\pi_0$, following the works \cite{ziegler2019fine, ouyang2022training}. Formally, regularized reward function is defined as: 
$\hat{R}(s_t, a_t, y) = R(s_t, a_t, y) - \lambda \text{KL} \left( \pi (a_t| s_t) || \pi_0 (a_t| s_t) \right)$ where $\text{KL} \left( \pi (a_t| s_t) || \pi_0 (a_t| s_t) \right) = (\log \pi (a_t| s_t) - \log \pi_0 (a_t| s_t))$ and $\lambda$ is the KL coefficient \cite{ouyang2022training}. Note we used use a fixed KL coefficient rather than an adaptive controller.

\subsection{Task Details}
\label{appendix:task}
\begin{table}[h]
    \centering
    \resizebox{\textwidth}{!}{
    \begin{tabular}{c|ccc}
        \toprule
        Task & Train/Val/Test & Prompt & Gen. Length\\
        \midrule
        \texttt{IMDB} & 25K/5K/5K & Partial movie review up to 64 tokens & 48\\
        \texttt{CommonGen} & 32651/993/1497 & "Generate a sentence with: " set of 3-5 concepts & 20\\
        \texttt{TL;DR} & 117000/6450/6550 & "TL;DR: " & 50\\
        \texttt{TL;DR Preference} & 92500/3300/8300 & "TL;DR: " & N/A\\
        \bottomrule
    \end{tabular}}
    \vspace{0.5mm}
    \caption{Train, val, test splits, prompts, and max generation length used for each task.}
    \label{tbl:dataset}
\end{table}

\paragraph{IMDB:} We experiment on the IMDB dataset for positive movie review generation. As shown in \Cref{tbl:dataset}, the dataset consists of 25k training, 5k validation and 5k test prompts of movie review text with either positive or negative sentiment labels. As in put to our models, we use partial movie reviews that are at most 64 tokens long and ask the model to complete the review with a positive sentiment with at most 48 generated tokens.

\paragraph{CommonGen:} CommonGen \cite{lin-etal-2020-commongen} is a common sense text generation task where the model is given a set of concepts (i.e. hockey, rink, game) and is asked to generate a semantically correct sentence using those concepts (i.e. the hockey team played a game at the rink). We follow the same splits as the dataset creators and refer the readers to Table 1 of \cite{lin-etal-2020-commongen} for more in-depth statistics of the dataset. In our experiments, we prompted out models with "\textit{generate a sentence with: }" and generated at most 20 tokens. We chose this generation length based on the maximum token length of the references in the training dataset.

\paragraph{TL;DR Summarization:} Following \cite{stiennon2020learning}, we evaluate on the summarization task. We use \url{CarperAI/openai_summarize_comparisons} for the preference reward training dataset and \url{CarperAI/openai_summarize_tldr} for the RL training dataset. For the SFT model that we use for our starting policy and our guide policy, we use the publicly available checkpoint \url{CarperAI/openai_summarize_tldr_sft}. We truncated/padded each prompt to 500 tokens on the GPT-J 6B tokenizer.

We first train our reward model using LoRA adapters. Our reward training is 1 epoch and where we got 70\% accuracy on the test set. With this reward model we run all of our experiments where our policy and critic are both LoRA adapters trained on top of SFT checkpoint.

\paragraph{Win Rate:} We calculated the win rate against the dataset references using Llama2-13B-chat \citep{touvron2023llama} publically available on HuggingFace. Following DPO \citep{rafailov2023dpo}, we prompt the model with instructions, 2 summaries (A) and (B), and instructions on how to answer. We randomize which summary is (A) or (B) when calculating the win rate over the test set. Below is our prompt skeleton:
\newpage
 \begin{verbatim}
  <<SYS>>
  You are an expert summary evaluator and can consistently
  distinguish between good and bad summaries. You provide
  informative, correct evaluations.
  <<\SYS>>

  Task: Judge the quality of two TLDRs, choose the options 
  among (A) or (B)
  context: [context]
  tldr (A): [summary 1]
  tldr (B): [summary 2]
  FIRST provide a one-sentence comparison of the two summaries,
  explaining which you prefer and why. SECOND, on a new line,
  state only (A) or (B) to indicate your choice. Your
  response should use hte format:
  Comparison: <one-sentence comparison and explanation>
  Preferred: <(A) or (B)>
 \end{verbatim}

\subsection{IMDB - Algorithm Details}
\Cref{tbl:imdb_hparams} lists the hyperparameters used in our IMDB experiments. Note that we used the same parameters here for all guide policies. Across all algorithms, we shared the same parameters as the ones we used for our \ppo{} baseline. Finally, we use top-k sampling with $K = 50$ as the decoding method and for fair comparison, we keep this setting for all methods.
\begin{table}[h]
\centering
   \begin{tabular}{ll}
       \toprule
       Setting & Values \\
       \midrule
       model & GPT2\\
       \midrule
       \ppo{} & steps per update: 1280 \\
              & total number of steps: 128000 \\
              & batch size: 64 \\
              & epochs per update: 5 \\
              & learning rate: 1e-6 \\
              & discount factor: 0.99 \\
              & gae lambda: 0.95 \\
              & clip ratio: 0.2 \\
              & value function coeff: 0.5\\
              & $\lambda$: 0.001 \\
              & $\eta$: 0.1 \\ 
       \midrule
       \cpi{} & Mixing Parameter ($\beta$): 0.2\\
       \midrule
       \agg{} & Mixing Parameter ($\beta$): 0.8\\
       \midrule
       \lol{} & Mixing Probability ($\alpha$): 0.8\\
       \midrule
       \lolsd{} & Stopping Time Iteration ($\alpha$): 20\\
       \midrule 
       decoding & sampling: true \\
                & top k: 50 \\
                & min length: 48 \\
                & max new tokens: 48\\
       \midrule
       tokenizer & padding side: left \\
                 & truncation side: left \\
                 & max length: 64\\
       \bottomrule
   \end{tabular}
   \vspace{0.5mm}
   \caption{Hyperparameters used for IMDB. Note that \cpi{}, \agg{}, \lol{}, and \lolsd{} all share the same PPO parameters. All processes use the same decoding and tokenizer parameters.}
   \label{tbl:imdb_hparams}
\end{table}
\newpage
\subsection{CommonGen - Algorithm Hyperparameters}
\begin{table}[htb]
\centering
   \begin{tabular}{ll}
       \toprule
       Setting & Values \\
       \midrule
       model & T5 \\
       \midrule
       \ppo{} & steps per update: 2560 \\
              & total number of steps: 1,280,000 \\
              & batch size: 640 \\
              & epochs per update: 4 \\
              & learning rate: Linear decay 1e-5 \\
              & discount factor: 0.99 \\
              & gae lambda: 0.95 \\
              & clip ratio: 0.2 \\
              & value function coeff: 30.0\\
              & $\lambda$: 0.001 \\
              & $\eta$: 0.1 \\ 
       \midrule
       \cpi{} & Mixing Parameter ($\beta$): 0.2\\
       \midrule
       \agg{} & Mixing Parameter ($\beta$): 0.8\\
       \midrule
       \lol{} & Mixing Probability ($\alpha$): 0.8\\
       \midrule
       \lolsd{} & Stopping Time Iteration ($\alpha$): 200\\
       \midrule 
       decoding & num beams: 5 \\
                & min length: 5 \\
                & max new tokens: 20 \\
       \midrule
       tokenizer & padding side: left \\
                 & max length: 20\\
       \bottomrule
   \end{tabular}
   \vspace{0.5mm}
   \caption{Hyperparameters used for CommonGen. Note that \cpi{}, \agg{}, \lol{}, and \lolsd{} all share the same PPO parameters. All processes use the same decoding and tokenizer parameters.}
   \label{tbl:commongen_hparams}
\end{table}
\Cref{tbl:commongen_hparams} lists the hyperparameters used in our CommonGen experiments. Note that we used the same parameters here for all guide policies. Across all algorithms, we shared the same parameters as the ones we used for our \ppo{} baseline. Finally, we use beam search with the number of beams = 5 as the decoding method for inference. Note that for training, we still used softmax sampling with default temperature. For fair comparison, we keep this setting for all methods. Finally, note that for CommonGen, we set the KL coefficient to 0.

\subsection{TL;DR Summarization - Algorithm Hyperparameters}
\begin{table}[htb]
\centering
   \begin{tabular}{ll}
       \toprule
       Setting & Values \\
       \midrule
       model & GPT-J \\
       \midrule
       \ppo{} & steps per update: 70,400 \\
              & total number of steps: 7,040,000 \\
              & batch size: 128 \\
              & epochs per update: 4 \\
              & learning rate: 1e-5 \\
              & discount factor: 1.0 \\
              & gae lambda: 0.95 \\
              & clip ratio: 0.2 \\
              & value function coeff: 0.2\\
              & $\lambda$: 0.002 \\
              & $\eta$: 0.1 \\ 
       \midrule
       \cpi{} & Mixing Parameter ($\beta$): 0.6\\
       \midrule
       \agg{} & Mixing Parameter ($\beta$): 1.0\\
       \midrule 
       decoding & max new tokens: 50 \\
       \midrule
       tokenizer & padding side: left \\
                 & truncation side: right \\
                 & max length: 50\\
       \midrule
       RM LoRA   & r: 32 \\
                 & $\alpha$: 32 \\
                 & dropout: 0.0\\
       \midrule
       Policy LoRA   & r: 8 \\
                     & $\alpha$: 64 \\
                     & dropout: 0.1\\
       \midrule
       Critic LoRA   & r: 8 \\
                     & $\alpha$: 64 \\
                     & dropout: 0.1\\
       \bottomrule
   \end{tabular}
   \vspace{0.5mm}
   \caption{Hyperparameters used for TL;DR summarization. Note that \cpi{}, \agg{} all share the same PPO parameters.}
   \label{tbl:tldr_hparams}
\end{table}
\Cref{tbl:tldr_hparams} lists the hyperparameters used in our TL;DR summarization experiments. Note that we used the same parameters here for all guide policies. Across all algorithms, we shared the same parameters as the ones we used for our \ppo{} baseline. The reward LoRA adapter was used to pretrain our model. During RL training, we kept the reward adapter frozen.

\section{IMDB Qualitative Examples}
\vspace{-3mm}
\begin{figure}[H]
\begin{lstlisting}
Sample 1
Prompt: not a complex character, but totally rotten. The relation between him and Gisburne is just hilarious. Actually, just looking at de Rainault sitting in his throne, bored, glaring, makes me laugh even before he has said anything. Another actor that deserves extra praise is Ray Winstone as Will Scarlet. You
GPT2: know, when you think he's a bad actor, and when you think the script is all about the bad actors. Scarlet is the typical actor with a weak, pathetic voice, who fails to show emotion or self-respect, and who
BC: will not be disappointed. The opening scenes are amazing and the final scene of the film is amazing. Every single minute of this film is a treat with every emotion the characters are experiencing. From the first frame of the
PPO: know he likes old movies. I gave this film a 3 and a half out of 10. And, of course, I don't blame you for falling asleep! I gave it a 10 because of my favorite
PPO++: see he seems only interested in money, which really makes the two seem like friends. One particularly funny scene involves Winstone and the rest of the cast in their studio for the first time since "Rampage". I absolutely LOVE the way
AggreVaTeD: will laugh heartily; he is brilliant and always entertaining. I also think the film is just wonderful. I love what Van Dyke is able to bring to the film. If you want to go great, I'm sure Van Dyke 
LOLS: great film. I absolutely loved his style and direction. One of my favorite movies of all time. The first 25 minutes are really good and really enjoyable. It's actually very touching and engaging. The cast, including James Gandolfini
D2LOLS: could never have come across this again. It was one of the funniest films I've ever seen, and it remains my favorite movie ever. I am absolutely honored and very excited to see it again!!! I really love it and look forward

Sample 2
Prompt: comes from Zekeria Ebrahimi, who plays the young Amir. It's not that it's bad, it's just that it's not as effective as it could be. He just can't translate the guilt that consumes Amir to the audience. In fact, I think I might have been a little lost at
GPT2: first. The movie has too much going for it, too many unanswered questions and some of the characters not getting the emotional payoff in a reasonable enough manner. It's nice to see a filmmaker who isn't a hater, but I feel
BC: the second half as it wasn't shown at all until the very end, at best. I liked the plot (which I think was very well done, considering it's a "bad cop story"). It's not
PPO: first. After listening to the great Ghazala, who's brilliant in her acting, I am very impressed with her acting performance. I loved how you have a young and very attractive girl named Zekeria who is portrayed quite well and
PPO++: last week's deadline. It's quite a clever and good film, as well as an entertaining one. 4/10. Recommended. 8/10
AggreVaTeD: first. The movie really deserves the credit - it really is a very moving story and a really moving movie that is an amazing film. I personally love Amir and think it's an extremely powerful film. It's one of the things that makes
LOLS: some points. As a whole it's pretty enjoyable and is very well-acted and well edited, and is quite a compelling, quirky and fun film. And just as important, I found some very well-acted, and well-acted
D2LOLS: first, when Amir realizes what's really going on. There are many terrific performances in this beautifully directed film. I can't say which is one of the best, which is certainly a shame because it's an enjoyable

Sample 3
Prompt: Well, now I finally HAVE seen the true garbage version - and it's Paul Schrader's DOMINION: PREQUEL TO THE EXORCIST. It was relentlessly talky, uninteresting, and insipid. Stellan Skarsgard's troubled priest was nowhere near as
GPT2: interesting as the other characters in his book - and most of the scenes in the book were just so poorly edited and over-the-top; the acting was too, horribly bad, and most of the dialogue was thrown out in a couple
BC: well developed as the novel and story had - but did bring some of the best bits to the screen. I watched the movie expecting a story - "The God" - but found it just an excuse for schlock
PPO: good (I guess), if anything, the sort of low-brow, uninteresting sort of horror that made an original film "The Cat With the Feeling" look like some sort of cheesy 80's horror flick from 1999.
PPO++: exciting. And in addition, the only really really interesting character in this movie was Paul Schrader, who is at his worst as a creepy, self-absorbed man in a bar. I can't say I've ever liked a
AggreVaTeD: original or entertaining as I'd anticipated, and in the end, I got to see some interesting and engaging dialog. I loved that scene with Schrader explaining how it would be fun for him to make a movie that would make people think twice
LOLS: good as his master. Still, the film's message is very much worth exploring. It was the perfect love story for me, as you would expect. I'm an evangelical and am an excellent fan of all things Christian. I love reading
D2LOLS: brilliant as the others, as well as beautifully acted. Of course, some points in the film are so well acted and acted that, despite all its flaws, this is a very touching film. The cinematography is
\end{lstlisting}
\caption{\textbf{Qualitiative Analysis of IMDB}}
\end{figure}
\section{CommonGen Qualitative Examples}
\label{appendix:commongen}
\begin{figure}[h]
\begin{lstlisting}
Sample 1
Prompt: generate a sentence with: dress perform stage.
T5: dress up and perform on stage
BC: A young girl in a black dress is performing on a stage.
PPO: performers dress up on the stage.
PPO++: a woman performing on the stage dressed up.
AggreVaTeD: young woman performing on stage in a black dress.
LOLS: people performing on the stage in a dressed up.
D2LOLS: A woman is performing on the stage in a dress.

Sample 2
Prompt: generate a sentence with: ball fall throw.
T5: ball falls off the ground
BC: A man throws a ball that has fallen to the ground.
PPO: the ball falling from the throw.
PPO++: the ball is thrown and the ball falls.
AggreVaTeD: A man is throwing the ball into the ground.
LOLS: throwing the ball after it fell from the sky.
D2LOLS: A man is throwing the ball after it fell.

Sample 3
Prompt: generate a sentence with: arm chest fold.
T5: arm folds in the chest
BC: He folds his arms over his chest, then he folds his arms over.
PPO: folded the arms in the chest.
PPO++: a man with his arms folded in the chest.
AggreVaTeD: folding his arm over his chest.
LOLS: A man folds his arms in the chest.
D2LOLS: A man with his arms folded in the chest.
\end{lstlisting}
\caption{\textbf{Qualitiative Analysis of CommonGen}}
\end{figure}

\clearpage
\newpage

\section{TL;DR Qualitative Examples}
\label{appendix:tldr}
\textcolor{red}{Content Warning: This section contains examples of harmful language.}
\begin{figure}[h]
\begin{lstlisting}
Sample 1
Prompt: SUBREDDIT: r/AskReddit
TITLE: Wife's "friend" is stealing money.  Any ideas on how to approach this?
POST: So my wife has a friend that I will call Julie.  My wife and Julie have been friends for about 6 years if I remember correctly.  Julie is a shit friend in my opinion and I have told the wife my feelings.  When they lived together Julie would start physical fights with my wife and spat on my wife's face once.  Now when Julie comes over to our house there is a 98%
Julie does not have any other good close friends in our town and her closest family is 5 hours away.  I am thinking that is why my wife doesn't get rid of Julie.  About 7 months ago we were all outside and Julie said that she needed to go inside to use the restroom.  As we soon outside I noticed that our bedroom light turned on, stayed on for less than a minute, and then went off.  
Later that night after Julie left when the wife and I went inside and the wife discovered that her purse had been moved and that she was missing $20.  I looked at my stuff and my checkbook had been moved.  The only other person in our house that night was Julie and she was left alone.
On Saturday my wife went shopping with Julie.  Again for some reason my wife left her purse with Julie alone.  Today my wife was trying to buy gas and is now missing $30.  
Before Julie started to steal from us her car battery died and she called us.  Stupidly I paid for and installed a new one for her.  I still haven't been paid back for that $87.
I am trying to figure out if it is time for me to say something to Julie or do I let the wife handle it?  
BC: Wife's "friend" is stealing money.  Any ideas on how to approach this? 
PPO: Wife's "friend" is stealing money.  Any ideas on how to approach this?  Wife's "friend" is a shit friend in my opinion and I have told the wife my feelings.
PPO++: Wife's "friend" is stealing money.  Any ideas on how to approach this?  Wife doesn't have any other good close friends in our town and her closest family is 5 hours away.
AggreVaTeD: Wife's "friend" is stealing money.  Any ideas on how to approach this?  Wife and I have lost $120 in the past 2 weeks.

Sample 2
Prompt: SUBREDDIT: r/Pets
TITLE: My sister's dog was picked up by animal control, $150 to get him out and she can't afford it.  He'll be euthanized in 2 days if she doesn't pay up.
POST: My sister has had this dog for the longest time, he's been shuffled around our family from time to time when she couldn't keep him due to being a renter.  I had him for a year and had to give him back to her because I moved into a smaller place.  I told her to keep him until the end of this year and I would take him back but he managed to get out of his yard today and got locked up by animal control it's going to cost her $150 to get him out or he'll be put to sleep!  He's about 11 years old, the sweetest beagle in the world! He doesn't bark he's great with kids and he's just a bug ol teddy bear! I don't know what to do!! I can't let him be put to sleep but I can't afford to help her! :-( 
BC:  My sister's dog was picked up by animal control, $150 to get him out and she can't afford it.  He'll be put to sleep if she doesn't pay up.
PPO: Sister's dog was picked up by animal control, $150 to get him out or he'll be put to sleep.  He's about 11 years old, the sweetest beagle in the world! He doesn't bark he's great with
PPO++: Sister's dog was picked up by animal control, $150 to get him out and she can't afford it.  He'll be euthanized in 2 days if she doesn't pay up.
AggreVaTeD:  sister's dog got locked up by animal control, $150 to get him out or he'll be put to sleep.  He's about 11 years old, the sweetest beagle in the world! He doesn't bark he's great with kids
\end{lstlisting}
\caption{\textbf{Qualitiative Analysis of TL;DR.}}
\end{figure}

\clearpage
\newpage

\end{document}